\documentclass{article}
\usepackage{spconf,amsmath,graphicx}
\usepackage{hyperref}       
\usepackage{url}            
\usepackage{booktabs}       
\usepackage{amsfonts}       
\usepackage{graphicx}
\usepackage{amsmath}
\usepackage{amssymb}
\usepackage{amsmath,bm}
\usepackage{nicefrac}       
\usepackage{microtype}      
\usepackage{subfigure}
\usepackage{placeins}
\usepackage{xcolor}

\title{Explainable AI for COVID-19 CT Classifiers: An Initial Comparison Study}

\name{Qinghao Ye $^{1,2}$, Jun Xia $^{3*}$ and Guang Yang $^{4,5}$\sthanks{Corresponding authors---J Xia: xiajun@email.szu.edu.cn \& G Yang: g.yang@imperial.ac.uk}}
\address{$^1$ University of California, San Diego, La Jolla, California, USA\\
	$^2$ Hangzhou Ocean's Smart Boya Co., Ltd\\
	$^3$ Radiology Department, Shenzhen Second People’s Hospital, Shenzhen, China\\
	$^4$ Royal Brompton Hospital, London, UK\\
	$^5$ National Heart and Lung Institute, Imperial College London, London, UK}
\begin{document}
%
\maketitle
\begin{abstract}

Artificial Intelligence (AI) has made leapfrogs in development across all the industrial sectors especially when deep learning has been introduced. Deep learning helps to learn the behaviour of an entity through methods of recognising and interpreting patterns. Despite its limitless potential, the mystery is how deep learning algorithms make a decision in the first place. Explainable AI (XAI) is the key to unlocking AI and the black-box for deep learning. XAI is an AI model that is programmed to explain its goals, logic, and decision making so that the end users can understand. The end users can be domain experts, regulatory agencies, managers and executive board members, data scientists, users that use AI, with or without awareness, or someone who is affected by the decisions of an AI model. Chest CT has emerged as a valuable tool for the clinical diagnostic and treatment management of the lung diseases associated with COVID-19. AI can support rapid evaluation of CT scans to differentiate COVID-19 findings from other lung diseases. However, how these AI tools or deep learning algorithms reach such a decision and which are the most influential features derived from these neural networks with typically deep layers are not clear. The aim of this study is to propose and develop XAI strategies for COVID-19 classification models with an investigation of comparison. The results demonstrate promising quantification and qualitative visualisations that can further enhance the clinician's understanding and decision making with more granular information from the results given by the learned XAI models. 

\end{abstract}

\section{Introduction}

Artificial intelligence, or AI, which refers to the creation of intelligent hardware or software that can emulate ``human'' behaviours such as learning and problem solving, is a broad branch of computer science that focuses on a machine's capability to generate rational behaviour. The goal of AI is to create systems that can perform tasks that would otherwise require human intelligence. AI manifests itself in everyday life through virtual assistants, search prediction technology, and driving automobiles. As AI multiplies in our daily lives, so does the fear that people will lose control. As a result, the European Commission, for example, has pushed forward to create a framework to regulate the use of AI, leading to guidelines set out the requirements that AI systems must meet to be trustworthy.

Deep Learning models such as deep neural networks are powerful AI models that can learn complex patterns from various types of data such as images, speech signals or text in order to predict the associated properties with impressive precision. However, many view these models as uninterpretable because the prediction patterns learned are difficult to interpret. This black-box nature has hindered the widespread application of deep neural networks to digital healthcare and medicine, where the interpretation of predictive patterns is of paramount importance. 

Despite the limitless potential of deep learning, the mystery is how the algorithms reach a decision in the first place. Reliability concerns are often raised in deep learning models, driving questions like, ``Which processes did you take over and at what speed? How did you make such an autonomous decision?''  While the deep learning model helps parse large amounts of data into intelligent insights for applications that range from fraud detection to weather forecasting, the human mind is constantly confused about how to reach such conclusions. Additionally, the recurring need to understand the processes behind decisions becomes increasingly important when the deep learning model has the potential to make decisions based on incomplete, error-prone, or biased information that important features may languish inside the black-box.

Explainable AI (XAI) is the key to unlocking AI and the black-box for deep learning. XAI is an AI model that is programmed to explain the goals, logic and decision making so that the common end user can understand it. Here, end users can be the designers of the AI systems or the people affected by the decisions of an AI model. According to Arrieta et al. \cite{arrieta2020explainable}, early AI systems were easy to interpret. For example, decision trees, Bayesian classifiers and other algorithms have a certain degree of traceability, visibility and transparency in their decision-making process. But recently, complex and opaque decision-making systems, such as deep neural networks, have appeared in AI. The empirical success of deep neural networks is based on a combination of efficient algorithms and their huge parametric space. This parametric space comprises hundreds of layers and millions of parameters, which means that deep neural networks are considered complex black-box models. The opposite of the black-box is the transparent white-box, which advocates the search for a direct understanding of the mechanism that how a model works. This demand is increasing also due to ethical concerns, for example, because the dataset used to train deep learning systems may not be justified, legitimate, or not allow detailed explanations of their behaviour. Additionally, for the opaque black-box AI decision making, XAI also addresses the bias inherent in AI systems. Bias in AI systems can have detrimental effects, especially in digital healthcare and medicine \cite{yang2021unbox, ye2019dual}. 

This study aims to build up a COVID-19 classification system that incorporates both the latest development of deep learning with the advances in XAI. In doing so, our data driven system can provide not only accurate classification outcomes but also explainable results for such a deep learning model.

\section{Related Work}

COVID-19 (a.k.a. Coronavirus 2019) has become a worldwide epidemic with rapid development and unprecedented spread of the disease since December 2019. The virus is highly contagious and can be asymptomatic and mostly latent, but it develops rapidly and can also lead to quickly progressive and frequently fatal pneumonia in 2–8\% of those affected \cite{harmon2020artificial}. However, the total mortality, prevalence, and severity of viral infection may be ill-defined partially because of the challenges associated with SARS-CoV-2 infection, such as an increase in viral load at symptoms or just beforehand, and misconception of multi-organ pathophysiology with dominant features and lung lethality. Differential proliferation complicates healthcare systems worldwide due to a lack of critical protective equipment and specialised providers, part of missing a cost-effective testing approach. When the fast reverse transcription polymerase chain reaction (RT-PCR) test becomes available, the challenges are with high false negative values, slow processing, inconsistency in test methods, and the detection sensitivity is often registered as low as 60–70 per cent \cite{hu2020weakly}.

Computed tomography (CT) is a test that offers a window to physiology that can explain the different stages of disease diagnosis and evolution. Although problems exist with the rapid detection of COVID-19, frontline radiologists find common features such as the ground glass opacity (GGO) in the lung periphery, rounded opacity, enlarged intra-infiltrate vessels, and further consolidations that are signs of developing a critical illness. Although CT and RT-PCR are most commonly concordant, CT may also detect early COVID-19 in patients with a negative RT-PCR examination, in patients without symptoms, before or after symptoms develop \cite{hu2020weakly}. The CT evaluation has been an important part of the initial examination of patients with suspected or confirmed COVID-19 in various centres in Wuhan China and northern Italy. A new international expert consensus study advocates the use of chest CT in COVID-19 patients with declining respiratory status or resource-restricted conditions for surgical triage of patients with moderate-severe clinical characteristics and a high pre-test likelihood of COVID-19 \cite{rubin2020role}. However, these guidelines also recommend against the use of chest CT in screening or diagnostic settings that may be partially due to similar radiographic presentation with other influenza-associated pneumonia. Techniques for distinguishing between these individuals can reinforce support for the use of CT in diagnostic settings.

Due to the dramatic rise in the number of newly suspected COVID-19 cases, the role of AI approaches in the identification or characterisation of COVID-19 imaging can be important \cite{roberts2021common}. CT offers a simple yet expeditious window for this process. Besides, deep learning of large-scale multinational CT data can provide automated and reproducible imaging biomarkers for the classification and quantification of COVID-19. Previous trials have shown the efficacy of AI for the diagnosis of COVID-19 infection or even distinguishing COVID-19 from community-acquired pneumonia. AI models are unavoidably restricted in clinical use due to the homogeneity of data sources, which in turn restricts their applicability to different populations, demographics, or geographies. Moreover, as mentioned, CT can share some common imaging properties between COVID-19 and other types of pneumonia and infection lesions may be confounded at the early stage, making automated classification difficult. More importantly, AI techniques, especially deep learning based methods, normally possess multi-layers of neural networks and the explanation of how the network components work synergistically and which are the most important features are commonly vague to the developers and end-users. This drives us to build up a robust deep neural network based model for distinguishing COVID-19 from CT scans with the assistance of both local and global XAI.

\section{Method}
In this section, we describe a classifier that is designed for distinguish OVID-19 from CT scan volumes with XAI solutions. Commonly, image-level labels are not easily obtained since annotating single images are labour-intensive and time-consuming, which takes a significant amount of time for well-trained radiologists to annotate them. On the other hands, patient-level annotations are available easily. To overcome these problems, we propose an explainable COVID-19 classifier for the acquired CT data. The XAI module in the proposed model can provide radiologists with auxiliary diagnostic information. Moreover, we also divide the images into super-pixels and adopt the LIME method \cite{ribeiro2016should} and compute the Sharply values for the sake of interpreting the individual contribution of each super-pixel and compare the local and global explainability. As a result, these XAI modules can be used to determine which part of the CT images contains the most predictive lesion areas leading the model to make the final decision.

\subsection{Explainable Module}
As widely used, Class Activation Map (CAM) \cite{zhou2016learning} can generate the localisation maps for the prediction through the combination of classification layer and feature maps of the backbone networks such as ResNet \cite{he2016deep}. Denoted the $d$-th feature maps of the network as $F^{d} \in \mathbb{R}^{H\times W}$, and the classification score $s_{c}$ for class $c$ can be obtained by
\begin{equation}
\label{eq:origin_cam}
    s_c = \sum_{d = 1}^{C'} W_{dc} \left( \frac{1}{HW} \sum_{i=1}^H \sum_{j=1}^W F^{d}_{ij} \right),
\end{equation}
where $W \in \mathbb{R}^{C'\times C}$ is the parameter of last classification layer. Therefore, the classification score $s_c$ can be computed as the weighted sum of feature maps after the Global Average Pooling (GAP) operation, and activation map $A_c$ for class $c$ can be computed as
\begin{equation}
    (A_c)_{ij} = \sum_{d = 1}^{C'}W_{dc}F^{d}_{ij}.
\end{equation}

\begin{figure}
    \centering
    \includegraphics[width=0.95\linewidth]{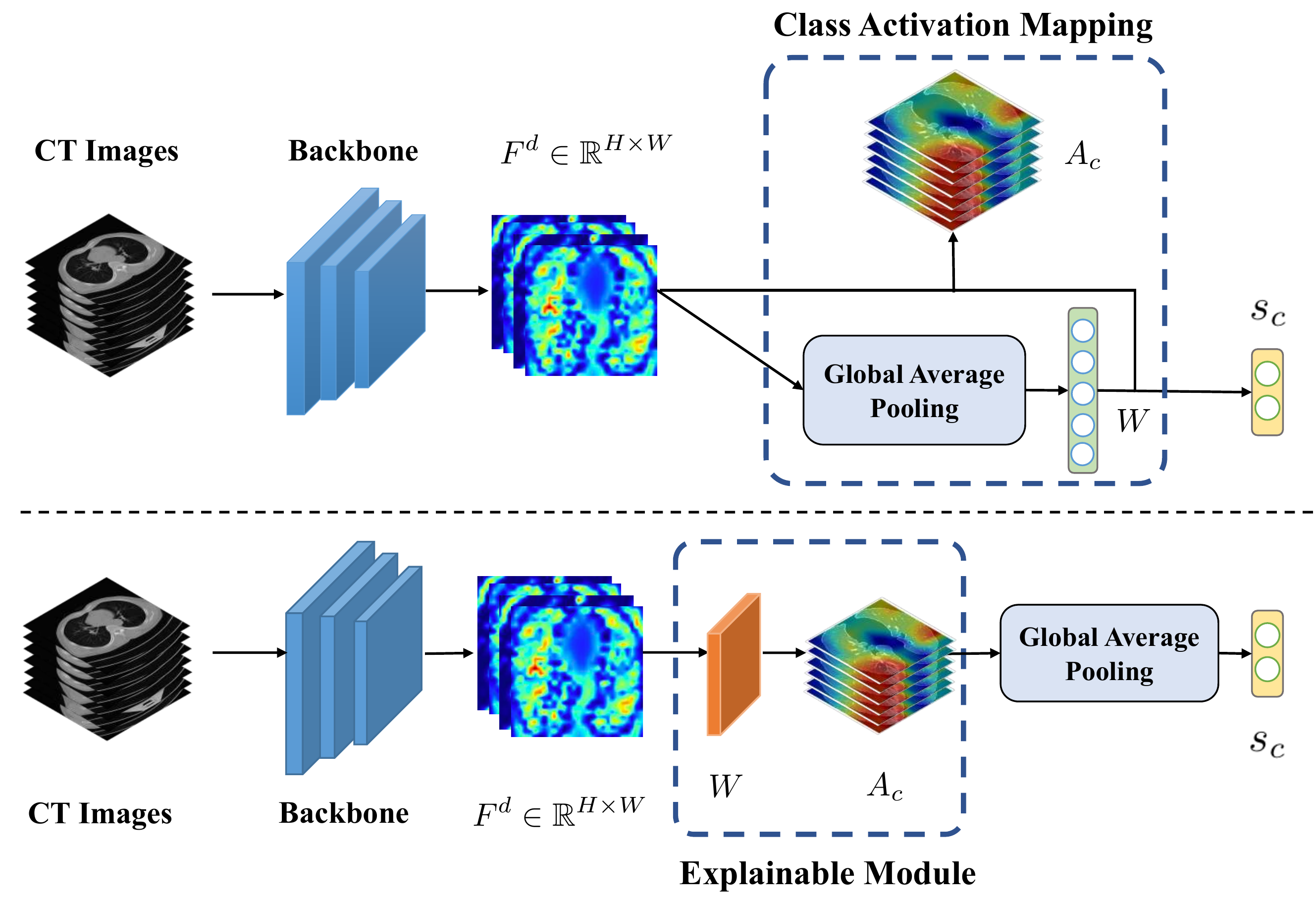}
\caption{The workflow comparison between class activation mapping and our proposed explainable module.}
\label{fig:arch_cam}
\end{figure}

However, generating the activation maps through CAM is a post-processing procedure, in which the network should be trained on the dataset then computes the feature maps and activation maps using the fixed weight $W$ from the classification layer leading to the extra computation. Besides, it cannot be visualised during the training procedure since the weight $W$ keeps updating. Therefore, to overcome this drawback, we replace the classification layer (fully connection layer) with a $1\times 1$ convolution layer since they are mathematically equivalent. Meanwhile, we reformulate the Equation (\ref{eq:origin_cam}) as
\begin{equation}
    s_c = \frac{1}{HW} \sum_{i=1}^H \sum_{j=1}^W \left(\sum_{d = 1}^{C'} W_{dc} F^{d}_{ij} \right) = \frac{1}{HW} \sum_{i=1}^H \sum_{j=1}^W (A_c)_{ij},
\end{equation}
where $A_c$ is the activation map for class $c$ that is learnt adaptively during the training procedure instead of obtained by post-processing. The workflow of CAM and our method are depicted in Figure \ref{fig:arch_cam}.

\begin{figure*}
    \centering
    \includegraphics[width=0.95\linewidth]{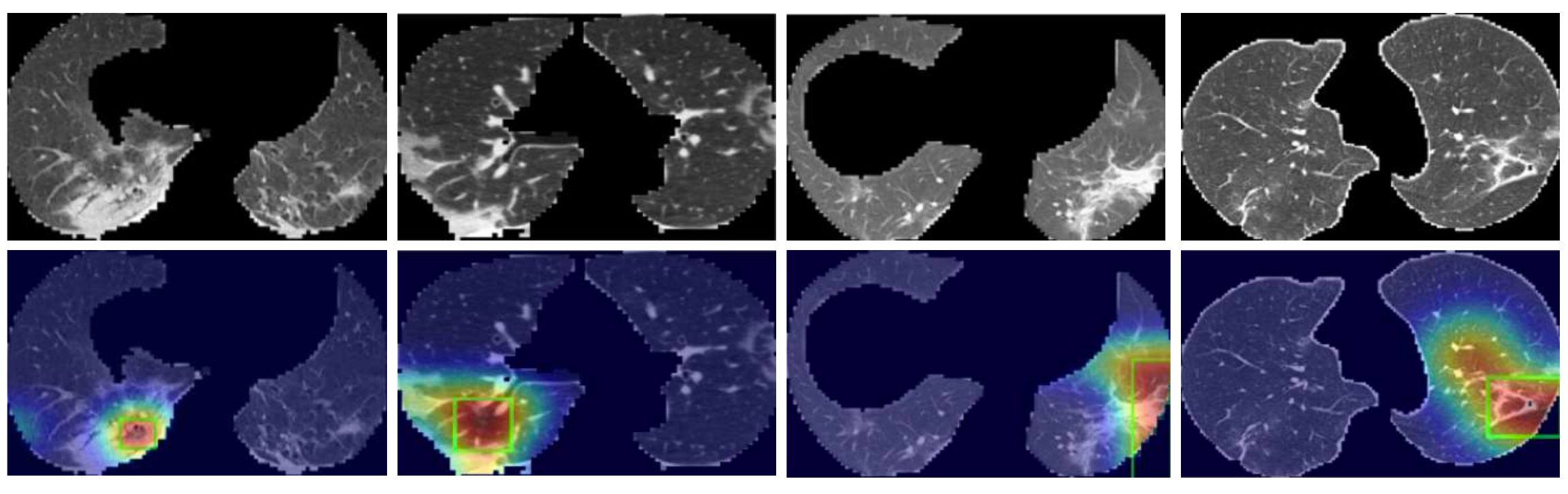}
\caption{Samples of the activation map $A_c$ generated by our proposed explainable module for classifying the COVID-19 positive cases. The first row contains the original input images, and the second row represents the corresponding activation maps and bounding boxes. The colour indicates the extent of the attention.}
\label{fig:cam}
\end{figure*}

\subsection{Slice Integration Module}
In intuition, the CT volumes of each patient has a different severity. Some of the COVID-19 positive patients have only few infections inside their lungs, which indicates most of the slices in the 3D volume are normal. In contrast, most of the patients with severe infections have larger lesions. Thus, the data would be extremely noisy when directly applying the patient-level annotation to each CT slice, which can affect the model performance. To avoid this problem, we develop a slice integration module, which integrates the probabilities of slices to reconstruct the probability of the whole CT volume. Besides, we first assume that the lesions are consecutive and consistently distributed in the CT volume. Under such assumption, we develop a section based strategy to handle this problem. Suppose that the patient $\mathcal{P} = [\mathcal{I}_1, \mathcal{I}_2, \cdots, \mathcal{I}_n]$ has $n$ CT slices, and we divide them into several disjoint sections $\mathcal{P} = \{S_i\}_{i=1}^{|S|}$ where $|S|$ is the number of sections defined by
\begin{equation}
    |S| = \max \left(1, \left\lfloor\frac{n}{l_s} \right\rfloor \right),
\end{equation}
where $l_s$ is the section length that we empirically set to 8. Then we compute the probability of the patient $\mathcal{P}$ by
\begin{equation}
    P(c | \mathcal{P}) = P(c | \{S_i\}_{i=1}^{|S|}) = 1 - \prod_{i=1}^{|S|} (1 - P(c | S_i)),
\end{equation}
where $P(c | S_i)$ is obtained by the average classification score of the largest $k$ slices with Sigmoid function. Then, patient annotation $y$ is used to compute the classification loss, which is formulated as
\begin{equation}
    \mathcal{L} = -\frac{1}{N} \sum_{i=1}^N \left[ y_i \log P(c = 1 | \mathcal{P}_i) + (1 - y_i) \log P(c = 0 | \mathcal{P}_i) \right].
\end{equation}

\subsection{Local Importance Explanation}
To explicitly highlight which local part of the slices can improve the model performance, we incorporate the LIME method \cite{ribeiro2016should} into our classifier. In details, the input slices are firstly divided into several super-pixels, which share similar visual properties among individual pixels within it. 
For trailing the influence of each super-pixel, each of them is randomly masked, and we define the interpretable model $g(x) = \sum_{i=1}^N w_i x_i$, where $N$ is number of super-pixels and $x_i = 1$ denotes the super-pixel is unmasked. Since we want to find the functionality of each super-pixel, the optimal interpretable model $\phi(x)$ is defined as
\begin{equation}
\label{eq:lime}
    \phi(x) = \mathop{\arg \min}_{g \in G} \sum_{(x, x')\in \mathcal{X}} e^{\frac{-(x - x')^2}{\sigma^2}} (f(x) - g(x'))^2 + \|x'\|_1,
\end{equation}
where $G$ is a candidate set of possible interpretable models, $f$ is the trained network with explainable module and slice integration module, $x$ is the original images, and $x'$ is the masked images. We empirically set $\sigma = 2$ for solving the optimal interpretable model. By applying the Lasso algorithm, we can solve the Equation \ref{eq:lime} for all of the weights $w_i$. Therefore, the super-pixels with positive weights would account for the prediction towards a specific class.

\subsection{Global Importance Explanation}
The LIME method indeed account for the contribution of each super-pixel with local explainability; however, it violates the convention that if the model changes so that the input's contribution increase, the input's attribution should not decrease. In other words, the model's interpretation should be consistency regardless of changing of the input. Therefore, SHapley Additive exPlanation (SHAP) value \cite{lundberg2017unified} is then incorporated for the measure of feature importance. It not only preserves the local accuracy of the super-pixels, but also follows the consistency assumption. Similarly, the SHAP value $w_i$ can be solved using a similar form of Equation (\ref{eq:lime}) as
\begin{equation}
\label{eq:shap}
    \phi(x) = \mathop{\arg \min}_{g \in G} \sum_{(x, x')\in \mathcal{X}} \frac{N - 1}{\|x\|_1 (N - \|x\|_1)} (f(x) - g(x'))^2.
\end{equation}
As a consequence, the SHAP values from game theory can be computed using the weighted linear regression. In specific, the region with negative SHAP value contributes negatively to the prediction, whereas the remainder of the feature have a positive contribution to the prediction.

\begin{table}
\begin{center}
\resizebox{1.0\linewidth}{!}{%
\begin{tabular}{l|c|c|c|c}
\hline
\textbf{Method}                      & \textbf{Accuracy (\%)} & \textbf{Precision (\%)} & \textbf{Recall (\%)} & \textbf{AUC (\%)} \\ \hline
 ResNet-50 \cite{he2016deep}        & 53.72                      & 61.42                   & 77.20                      & 46.30       \\
COVID-Net \cite{wang2020deep}       & 53.65                      & 61.35                   & 77.20                     & 44.53      \\
COVNet \cite{li2020artificial}       & 67.60                      & 76.01                   & 73.18                      & 66.13     \\
VB-Net \cite{ouyang2020dual}         & 76.73                      & 85.26                   & 77.58                      & 89.48      \\
Ours                      & \textbf{89.23}             & \textbf{89.98}          & \textbf{93.86}          & \textbf{93.22}      \\ \hline
\end{tabular}
}
\end{center}
\caption{Comparison results of our method vs. state-of-the-art architectures on the CC-CCII dataset.}
\label{tb:sota}
\end{table}

\section{Experiments}
\subsection{Datasets}
We collected CT volume data from four hospitals in China and removed the personal sensitive information to ensure the privacy. In total, there were 380 CT volumes with COVID-19 positive, and 424 COVID-19 negative CT volumes. For a fair comparison, we trained the models on these private datasets, then tested on the open access CC-CCII dataset \cite{zhang2020clinically}, which is a publicly available dataset with 2,034 CT volumes.

\subsection{Implementation Details}
Following the protocol used in \cite{zhang2020clinically}, we firstly utilised the U-Net to segment the CT images, and we randomly cropped the volumes and resized each image into $224 \times 224$. Meanwhile, the images in the input volume were randomly flipped horizontally with a probability of 0.5. We trained the models for 200 epochs with the learning rate of 0.001.

\subsection{Quantitative Results}
For comparison, we selected several state-of-the-art methods \cite{he2016deep, wang2020deep, li2020artificial, ouyang2020dual} that worked well on COVID-19 image classification. For those only worked on image-level methods \cite{he2016deep, wang2020deep}, we directly applied the patient-level annotations as image labels to train the network. In particular, VBNet \cite{ouyang2020dual} utilised the 3D residual convolutional blocks to train the network on the whole CT volumes rather than processing each CT slice. Meanwhile, COVNet \cite{li2020artificial} took the largest probability among the those of images in the volume as the volume probability. 

The results are summarised in Table \ref{tb:sota}. From the table, we can observe that our method outperformed other methods by a large margin. In particular, our method surpassed VBNet on accuracy by 12.5\%, which means that our method could be practical for real-world scenario. The slice integration module in our method could weight the slices in each section while preserving the characteristics of each section, leading to better accuracy. Besides, models (i.e., ResNet-50 and COVID-Net) trained on single images cannot achieve desirable results due to a large amount of noise in the training labels. Though COVNet took the most discriminative slice as the representation of the whole volume, it would perform poorly on negative examples since it brings bias towards positive cases.

\subsection{Qualitative Results}

Apart from robust quantitative performance of our model, the explainability of the model is also promising. To make the prediction to be more explainable, we firstly extracted the activation maps generated by explainable module of our method as described above and visualised in Figure \ref{fig:cam} . It can be observed that our proposed method would focus on the most discriminative part of the image, which can be identified as the lesion area, and make the positive prediction. Therefore, the radiologists can make auxiliary judgement based on the results provided by the activation maps. Meanwhile, the lesion bounding boxes can be extracted by our method, which tends to yield on the lesion parts of the generated activation maps, which further verified that our method was applicable to be a valuable auxiliary diagnosis tool for radiologists.

\begin{figure}
    \centering
    \includegraphics[width=0.95\linewidth]{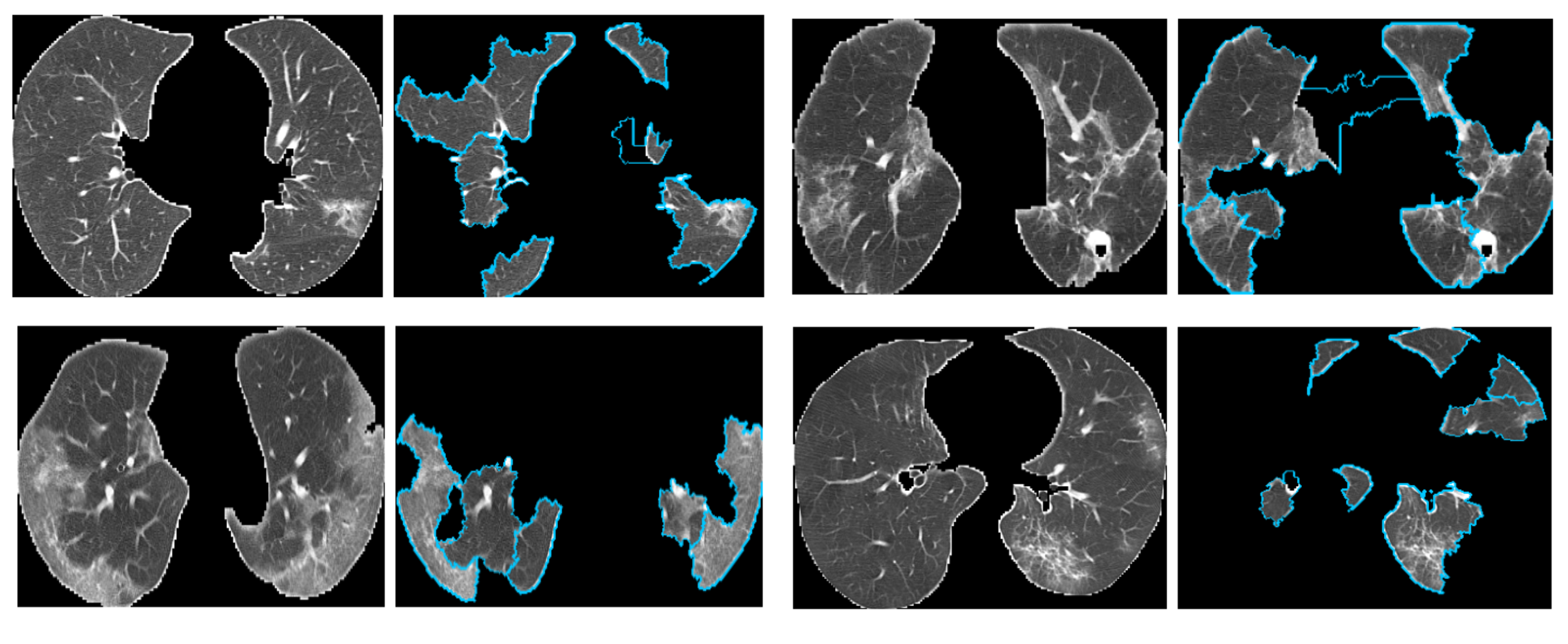}
\caption{Visualisation of the super-pixels contributed positively to the model prediction through the developed LIME paradigm.}
\label{fig:lime}
\vspace{-0.3in}
\end{figure}

Besides, instead of only showing the most salient parts of the image, we also focus on local regional contribution to the prediction. As we described above, we divided the input images into several super-pixels, in which pixels shared the similar visual pattern. Figure \ref{eq:lime} provides the examples of explaining the COVID-19 images from the positive CT volumes via the developed LIME method. For each pair of images, we highlighted those super-pixels that contributed positively to the prediction, and we can clearly observed that those regions contains lesions are highlighted for the prediction. Therefore, it demonstrates that the infected areas are easily to be spotted by this method.

\begin{figure}
    \centering
    \includegraphics[width=1\linewidth]{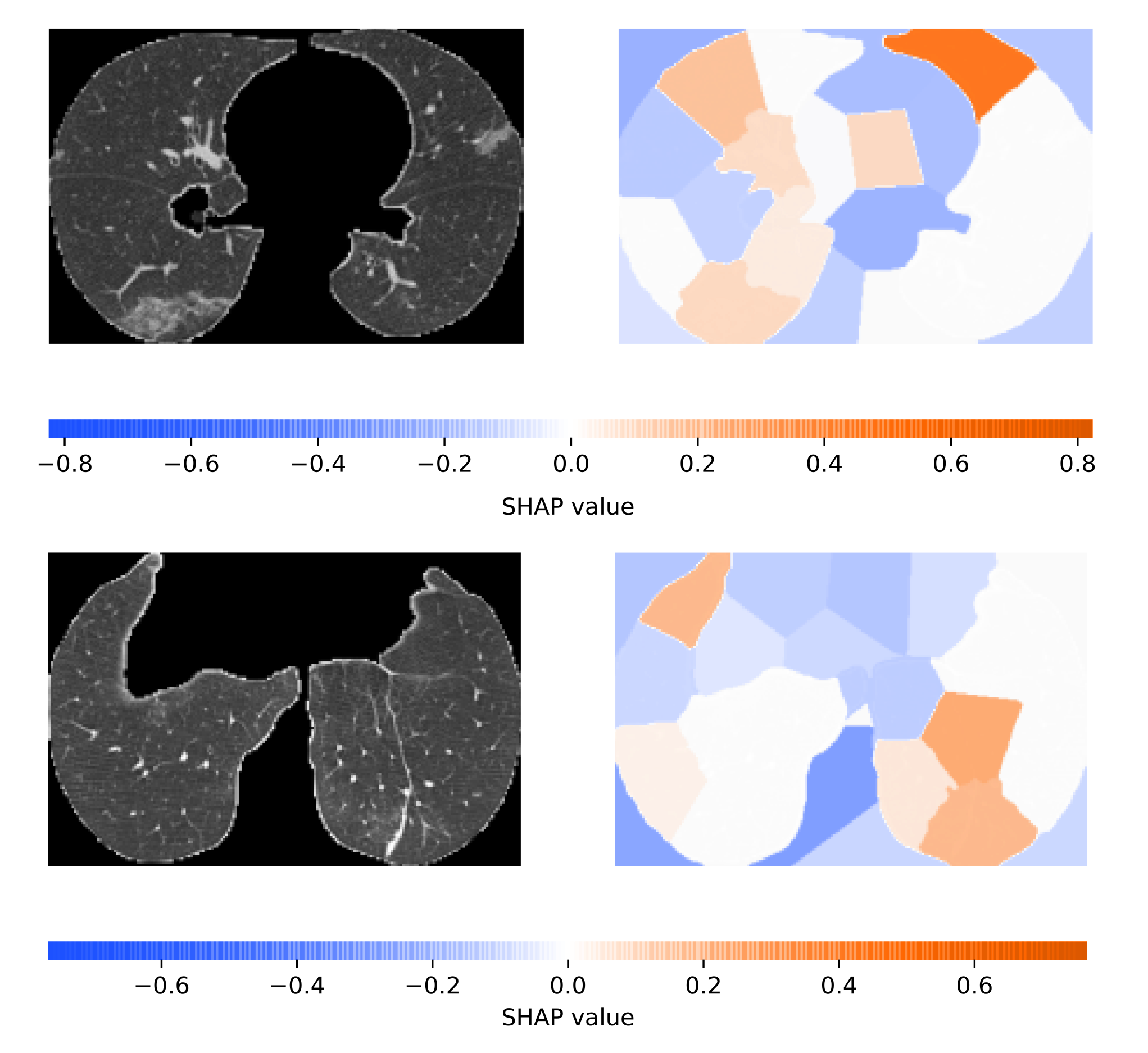}
\caption{Visualisation of the SHAP values for each super-pixel. The positive value indicates how much the corresponding super-pixel would encourage the network to make a positive prediction.}
\label{fig:shap}
\vspace{-0.3in}
\end{figure}

Furthermore, to overcome the shortcoming of the LIME method, we also visualised the contribution of each super-pixel. Figure \ref{fig:shap} provides two examples of COVID-19 positive predictions. We can find that the background in the images tend to negatively contribute to the prediction which are regard as the noise in the images. Besides, most of the lesion areas in the CT images obtain the large positive SHAP values indicating they significantly positively contribute to the final positive prediction, which are able to explain the prediction of the model.

\section{Conclusion}

In this study, we have presented an XAI equipped classifier for COVID-19 detection from CT images. The experimental studies have demonstrated that our XAI enhanced classifier model can not only provide promising and robust classification results but can also provide convincing explanation of these results. We have also compared the widely used CAM, LIME and SHAP value based XAI modules. Although the study has been carried out for COVID-19 studies, the application on other image modalities and clinical questions can be envisioned. 

\bibliographystyle{IEEEbib}
\bibliography{strings,refs}

\begin{thebibliography}{10}

\bibitem{arrieta2020explainable}
Alejandro~Barredo Arrieta, Natalia D{\'\i}az-Rodr{\'\i}guez, Javier Del~Ser,
  Adrien Bennetot, Siham Tabik, Alberto Barbado, Salvador Garc{\'\i}a, Sergio
  Gil-L{\'o}pez, Daniel Molina, Richard Benjamins, et~al.,
\newblock ``Explainable artificial intelligence (xai): Concepts, taxonomies,
  opportunities and challenges toward responsible ai,''
\newblock {\em Information Fusion}, vol. 58, pp. 82--115, 2020.

\bibitem{yang2021unbox}
Guang Yang, Qinghao Ye, and Jun Xia,
\newblock ``Unbox the black-box for the medical explainable ai via multi-modal
  and multi-centre data fusion: A mini-review, two showcases and beyond,''
\newblock {\em arXiv preprint arXiv:2102.01998}, 2021.

\bibitem{ye2019dual}
Qinghao Ye, Daijian Tu, Feiwei Qin, Zizhao Wu, Yong Peng, and Shuying Shen,
\newblock ``Dual attention based fine-grained leukocyte recognition for
  imbalanced microscopic images,''
\newblock {\em Journal of Intelligent \& Fuzzy Systems}, vol. 37, no. 5, pp.
  6971--6982, 2019.

\bibitem{harmon2020artificial}
Stephanie~A Harmon, Thomas~H Sanford, Sheng Xu, Evrim~B Turkbey, Holger Roth,
  Ziyue Xu, Dong Yang, Andriy Myronenko, Victoria Anderson, Amel Amalou,
  et~al.,
\newblock ``Artificial intelligence for the detection of covid-19 pneumonia on
  chest ct using multinational datasets,''
\newblock {\em Nature communications}, vol. 11, no. 1, pp. 1--7, 2020.

\bibitem{hu2020weakly}
Shaoping Hu, Yuan Gao, Zhangming Niu, Yinghui Jiang, Lao Li, Xianglu Xiao,
  Minhao Wang, Evandro~Fei Fang, Wade Menpes-Smith, Jun Xia, et~al.,
\newblock ``Weakly supervised deep learning for covid-19 infection detection
  and classification from ct images,''
\newblock {\em IEEE Access}, vol. 8, pp. 118869--118883, 2020.

\bibitem{rubin2020role}
Geoffrey~D Rubin, Christopher~J Ryerson, Linda~B Haramati, Nicola Sverzellati,
  Jeffrey~P Kanne, Suhail Raoof, Neil~W Schluger, Annalisa Volpi, Jae-Joon Yim,
  Ian~BK Martin, et~al.,
\newblock ``The role of chest imaging in patient management during the covid-19
  pandemic: a multinational consensus statement from the fleischner society,''
\newblock {\em Chest}, vol. 158, no. 1, pp. 106--116, 2020.

\bibitem{roberts2021common}
Michael Roberts, Derek Driggs, Matthew Thorpe, Julian Gilbey, Michael Yeung,
  Stephan Ursprung, Angelica~I Aviles-Rivero, Christian Etmann, Cathal McCague,
  Lucian Beer, et~al.,
\newblock ``Common pitfalls and recommendations for using machine learning to
  detect and prognosticate for covid-19 using chest radiographs and ct scans,''
\newblock {\em Nature Machine Intelligence}, 2021.

\bibitem{ribeiro2016should}
Marco~Tulio Ribeiro, Sameer Singh, and Carlos Guestrin,
\newblock ``{``Why should I trust you?'' Explaining the predictions of any
  classifier},''
\newblock in {\em Proceedings of the 22nd ACM SIGKDD international conference
  on knowledge discovery and data mining}, 2016, pp. 1135--1144.

\bibitem{zhou2016learning}
Bolei Zhou, Aditya Khosla, Agata Lapedriza, Aude Oliva, and Antonio Torralba,
\newblock ``Learning deep features for discriminative localization,''
\newblock in {\em Proceedings of the IEEE conference on computer vision and
  pattern recognition}, 2016, pp. 2921--2929.

\bibitem{he2016deep}
Kaiming He, Xiangyu Zhang, Shaoqing Ren, and Jian Sun,
\newblock ``Deep residual learning for image recognition,''
\newblock in {\em Proceedings of the IEEE conference on computer vision and
  pattern recognition}, 2016, pp. 770--778.

\bibitem{lundberg2017unified}
Scott~M. Lundberg and Su{-}In Lee,
\newblock ``A unified approach to interpreting model predictions,''
\newblock in {\em Advances in Neural Information Processing Systems 30: Annual
  Conference on Neural Information Processing Systems}, Isabelle Guyon, Ulrike
  von Luxburg, Samy Bengio, Hanna~M. Wallach, Rob Fergus, S.~V.~N.
  Vishwanathan, and Roman Garnett, Eds., 2017, pp. 4765--4774.

\bibitem{wang2020deep}
Shuai Wang, Bo~Kang, Jinlu Ma, Xianjun Zeng, Mingming Xiao, Jia Guo, Mengjiao
  Cai, Jingyi Yang, Yaodong Li, Xiangfei Meng, et~al.,
\newblock ``A deep learning algorithm using ct images to screen for corona
  virus disease (covid-19),''
\newblock {\em MedRxiv}, 2020.

\bibitem{li2020artificial}
Lin Li, Lixin Qin, Zeguo Xu, Youbing Yin, Xin Wang, Bin Kong, Junjie Bai,
  Yi~Lu, Zhenghan Fang, Qi~Song, et~al.,
\newblock ``Artificial intelligence distinguishes covid-19 from community
  acquired pneumonia on chest ct,''
\newblock {\em Radiology}, p. 200905, 2020.

\bibitem{ouyang2020dual}
Xi~Ouyang, Jiayu Huo, Liming Xia, Fei Shan, Jun Liu, Zhanhao Mo, Fuhua Yan,
  Zhongxiang Ding, Qi~Yang, Bin Song, et~al.,
\newblock ``Dual-sampling attention network for diagnosis of covid-19 from
  community acquired pneumonia,''
\newblock {\em arXiv preprint arXiv:2005.02690}, 2020.

\bibitem{zhang2020clinically}
Kang Zhang, Xiaohong Liu, Jun Shen, Zhihuan Li, Ye~Sang, Xingwang Wu, Yunfei
  Zha, Wenhua Liang, Chengdi Wang, Ke~Wang, et~al.,
\newblock ``Clinically applicable ai system for accurate diagnosis,
  quantitative measurements, and prognosis of covid-19 pneumonia using computed
  tomography,''
\newblock {\em Cell}, vol. 181, no. 6, pp. 1423--1433, 2020.

\end{thebibliography}

\end{document}